# PET Tracer Conversion among Brain PET via Variable Augmented Invertible Network

Bohui Shen, Wei Zhang, Xubiao Liu, Pengfei Yu, Shirui Jiang, Xinchong Shi, Xiangsong Zhang, Xiaoyu Zhou, Weirui Zhang, Bingxuan Li, Qiegen Liu, *Senior Member, IEEE*

*Abstract*—Positron emission tomography (PET) serves as an essential tool for diagnosis of encephalopathy and brain science research. However, it suffers from the limited choice of tracers. Nowadays, with the wide application of PET imaging in neuropsychiatric treatment, 6-$^{18}$F-fluoro-3, 4-dihydroxy-L-phenylalanine (DOPA) has been found to be more effective than $^{18}$F-labeled fluorine-2-deoxyglucose (FDG) in the field. Nevertheless, due to the complexity of its preparation and other limitations, DOPA is far less widely used than FDG. To address this issue, a tracer conversion invertible neural network (TC-INN) for image projection is developed to map FDG images to DOPA images through deep learning. More diagnostic information is obtained by generating PET images from FDG to DOPA. Specifically, the proposed TC-INN consists of two separate phases, one for training traceable data, the other for rebuilding new data. The reference DOPA PET image is used as a learning target for the corresponding network during the training process of tracer conversion. Meanwhile, the invertible network iteratively estimates the resultant DOPA PET data and compares it to the reference DOPA PET data. Notably, the reversible model employs variable enhancement technique to achieve better power generation. Moreover, image registration needs to be performed before training due to the angular deviation of the acquired FDG and DOPA data information. Experimental results exhibited excellent generation capability in mapping between FDG and DOPA, suggesting that PET tracer conversion has great potential in the case of limited tracer applications.

This work was supported in part by Chinese National Natural Science Foundation (82271267, 62201193), and the Natural Science Foundation of Guangdong Province (2022A1515011670). (Corresponding authors: B. Li and Q. Liu.) (B. Shen and W. Zhang are co-first authors.)

B. Shen, W. Zhang, X. Liu and Q. Liu are with School of Information Engineering, Nanchang University, Nanchang 330031, China. ({416100210118, 416100220076, zhangwei1}@email.ncu.edu.cn, liuqiegen@ncu.edu.cn)

P. Yu is with School of Mathematics and Computer Sciences, Nanchang University, Nanchang 330031, China. (419100220034@email.ncu.edu.cn)

S. Jiang is with School of Neurology, South China Hospital, Shenzhen University, Shenzhen, China. (JsrNeurology@outlook.com)

X. Shi and X. Zhang are with School of Nuclear, The First affiliated hospital of Sun Yat-Sen University, Guangzhou 510080, China. ({shixch, zhxiangs}@mail.sysu.edu.cn)

X. Zhou and W. Zhang are with School of Biomedical Engineering, Huazhong University of Science and Technology, Wuhan 430074, China. ({xiaoyunzhou, weiruizhang}@hust.edu.cn)

B. Li is with Institute of Artificial Intelligence, Hefei Comprehensive National Science Center, Hefei 230088, China. (libingxuan@iai.ustc.edu.cn)

## I. INTRODUCTION

Positron emission tomography (PET), as a non-invasive molecular imaging technique that generates images of humans and animals with high biochemical sensitivity. It is widely used for clinical research in oncology, neurology and cardiology [1]-[5]. It usually uses positron emitter-labeled glucose, amino acids, choline, thymine, receptor ligand and blood flow imaging agents as tracers. By displaying the state of metabolism, function, blood flow, cell proliferation and receptor distribution of the organism and focal tissues at the molecular level, it can provide more physiological and pathological diagnostic information to the clinic.

As more positron isotopes are manufactured, tracers are gradually developed and applied. Based on the success of $^{14}$C-labeled substances (e.g., $^{14}$C deoxyribose) for quantitative autoradiography experiments in the brain, PET with $^{11}$C- and $^{18}$F-labeled compounds were introduced in the late 1970s [6]-[10]. Since then, a number of different $^{11}$C and $^{18}$F labeled tracers have been designed for using in individual research projects and for relatively limited clinical applications. Afterwards, especially with the appearance of $^{18}$F-labeled fluoro-2-deoxyglucose ($^{18}$F-FDG), PET quickly became a major diagnostic standard of oncology. For instance, Ell *et al.* evaluated the response of $^{18}$F-FDG PET in cancer treatment. They suggested a major clinical need for $^{18}$F-FDG PET, in large part because of the inherent ability of $^{18}$F-FDG PET (before other response markers) to demonstrate whether disease changes have occurred or not [11]. After that, Omami *et al.* presented some basic fundamentals and applications of $^{18}$F-FDG PET in oral cavity and maxillofacial imaging. They presented a set of principles for $^{18}$F-FDG PET in the evaluation and management program for oral cavity and oropharyngeal squamous cell carcinoma, with pictorial illustrations and clinical applications [12]. However, a major obstacle to further expand the use of PET is the lack of approved licensed radiopharmaceuticals other than $^{18}$F-FDG [13].

In addition to $^{18}$F-FDG, another tracer 6-$^{18}$F-fluoro-3,4-dihydroxy-L-phenylalanine ($^{18}$F-DOPA), is gradually being discovered and used by researchers as an amino acid analogue for PET imaging. It has also been shown to be a good tracer for neuropsychiatric diseases [14]. As early as in 1996, Morrish *et al.* demonstrated that $^{18}$F-DOPA PET can be used to study disease progression. Assessment of Putamen function alone results in a good compromise between reproducibility and sensitivity [15]. It largely suggests that $^{18}$F-DOPA provides an objective measure of the functional integrity of dopaminergic terminals in the basal ganglia, which can be used prospectively to study the progression of the disease. Despite the general applicability of $^{18}$F-DOPA imaging in nuclear medicine, $^{18}$F-DOPA is available in a few countries worldwide. The main drawback that limits the widespread use of DOPA is its labeling method. Fortunately, Eggers et al. [16] compared the clinical and imaging findings of Parkinson's disease (PD) subtypes when studying PD, aiming to establish the relationship between clinical subtypes and dopamine and glucose metabolism. They demonstrated a direct interaction of striatal dopaminergic and glucose metabolism in different subtypes of PD. This study also demonstrated for the first time, consistent reductions in dopaminergic and glucose metabolism in the anterior striatum of patients. On the other hand, Kaasinen et al. [17] showed that principal component analysis revealed a positive correlation between the major components of $^{18}$F-DOPA uptake (associated with striatum uptake) and $^{18}$F-FDG uptake (thalamus and cerebellum with positive loading). These results indicate that there is a certain transformation relationship between different tracers used in PD diagnostic.

Recent years, the rapid expansion of deep learning in both industry and academia have inspired many research groups to integrate deep learning-based methods into medical imaging and radiation therapy [18]. The emergence of end-to-end network has made it possible for researchers to learn the mapping relationship, transformation and generation between different data. For example, Dong et al. [19] proposed an end-to-end mapping between low/high-resolution images by using convolutional neural networks (CNNs). Nevertheless, CNN-based synthesis measures still have a potential loophole in that contextual information can easily be ignored when synthesizing images. In addition, generative adversarial network (GAN) has been widely applied in the field of image synthesis [20]-[22]. In 2017, a method of learning how to translate images from source domain to target domain without paired examples by using CNN network was presented by Zhu et al. [21]. Later, Isola et al. [22] investigated conditional adversarial networks as a general-purpose solution to image-to-image translation problem. Recently, several variants or hybrid approaches that taking advantage of auto-encoders (AE) and GAN have been developed [23]-[26]. For example, Sharma et al. [24] and Zhou et al. [25] investigated variants of GAN to leverage implicit condition in multi-modal to achieve multiple-to-one mapping. Furthermore, Sun et al. [26] reported a flow-based generative model with AE-like network architecture for MRI-to-PET image generation. Although GAN exhibits strong generative ability, the reverse problem remains ambiguous. It's worth noting that, due to the invertibility of Invertible Neural Networks (INNs), the corresponding inverse process model is learned implicitly.

Importantly, different amounts and types of tracers for diseases with different sensitivities are used in these PET pathology work. Different tracers also reflect different focal areas. Although the excellent performance of $^{18}$F-DOPA in the brain region has made it a place in the tracer field [17], different from $^{18}$F-FDG, $^{18}$F-DOPA is not widely used around the world. Therefore, finding the mapping relationship between the information displayed by $^{18}$F-FDG and the information displayed by $^{18}$F-DOPA for PD diagnosis can further solve the problem that $^{18}$F-DOPA is not widely used as $^{18}$F-FDG, thus improving the diagnostic efficiency of PD. Inspired by [27-28], this work develops and implements a deep learning method based on reversible network (TC-INN) for learning the mapping relationship between $^{18}$F-FDG PET and $^{18}$F-DOPA PET, aiming to generate $^{18}$F-DOPA PET brain lesion maps from $^{18}$F-FDG PET brain lesion maps.

The main contributions of this work are summarized as follows:
- In order to solve the issue of tracer mapping and make DOPA-like diagnostics technique applied in more areas, this work first introduces invertible networks in the unified framework to implement the conversion of $^{18}$F-FDG brain image to $^{18}$F-DOPA brain image. Subject to the limitations of reversible network requirements, then we register FDG image and DOPA image before training. In addition, we utilize channel-copy technology to enhance the generation capability of reversible network.
- To the best of our knowledge, it is the first study to utilize DOPA-PET data from a digital PET instrument. In this work, mapping between different tracer images is realized for the first time. Compared with conventional PET imaging, digital PET has higher resolution brain imaging and therefore higher conversion accuracy between FDG and DOPA. In addition, it can provides richer diagnostic information for acquiring data from one tracer.

## II. PRELIMINARY

Several researchers have found that traditional convolutional neural networks lose some important depth information in the input image [29]. To fill this gap, INNs have been introduced that are capable of learning reversible representations to reduce information loss under specific conditions [30], and inverse scenes modeling [31]. INNs are characterized by three properties: First, the mapping from input to output is bijective, i.e., there is an inverse mapping. Second, both forward and inverse mappings are efficiently computable. Finally, both mappings have a tractable Jacobian that allows for the explicit computation of a posteriori probabilities.

INNs are a class of networks that provide bijective mappings between input and output. The general diagram of INNs is illustrated in Fig. 1. INNs learn mappings as $y = f(x)$, and fully invertible mappings as $x = f^{-1}(y)$. For an image pair $(x_i, y_i)$, with only one invertible network $f$, inputting the image $x_i$ into the network yields the output image $y_i$ and vice versa. Thus, the information is fully preserved for both forward and reverse transformations.

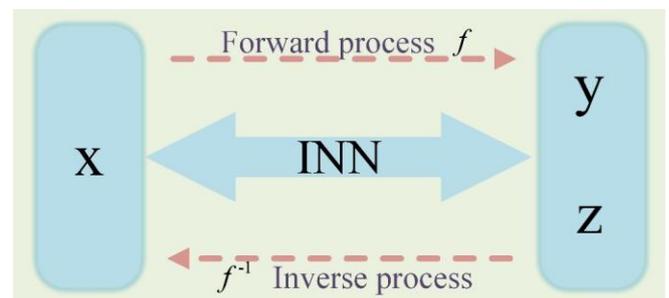

Fig. 1. The diagram of INN includes forward and inverse processes. The mapping from inputs to outputs is bijective, i.e., its inverse exists.

A typical design of INN contains a serial sequence of invertible blocks, each consists of two coupled layers [23]. We define $u$ as every block's input vector (for example, $u$ can be the model vector), which is split into two halves $u_1$ and $u_2$. They are transformed by an affine function via coefficients $\exp(s_i)$ and $t_i$ to produce the output $[v_1, v_2]$:

$$v_1 = u_1 \odot \exp(s_2(u_2)) + t_2(u_2) \quad (1)$$
$$v_2 = u_2 \odot \exp(s_1(u_1)) + t_1(u_1) \quad (2)$$

where $\odot$ represents element-wise multiplication. This process is trivially invertible for any functions $t$ and $s$:

$$u_2 = (v_2 - t_1(u_1)) \odot \exp(-s_1(u_1)) \quad (3)$$
$$u_1 = (v_1 - t_2(u_2)) \odot \exp(-s_2(u_2)) \quad (4)$$

Importantly, functions $s_i$ and $t_i$ do not need to be invertible themselves. Fig. 2 details the forward and inverse structure of each reversible block.

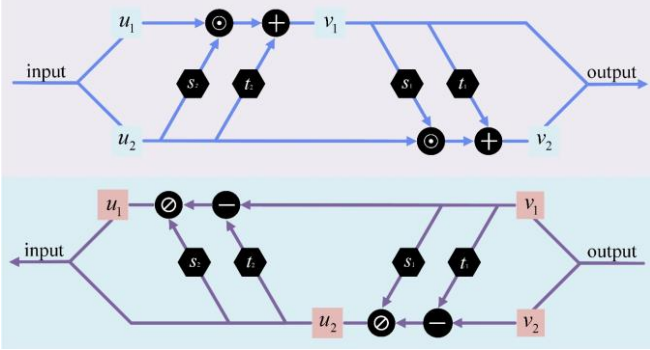

**Fig. 2.** The forward and inverse structure for each invertible block.

## III. METHOD

### A. Overview of TC-INN

To solve the tracer mapping issue and provide DOPA-like diagnostic method, this work proposes a deep learning algorithm TC-INN based on invertible network to achieve the conversion from FDG to DOPA. The algorithm only learns potential correlations between modes without disturbing the mode-specific structure. In addition, it provides an invertible connection between the original image and the target image. The visualization of the training process in TC-INN exhibited in Fig. 3. The forward process of TC-INN generates the synthesized images and the inverse process aims to recover the original images. However, it's worth noting that this work focuses more on learning the difference between input and output to achieve input-to-output generation. The combination of forward loss and backward loss is only used to optimize TC-INN. Specifically, the training data of the proposed invertible network consists of $^{18}$F-FDG PET images as input (original data). Subsequently, the estimated output data through the network is compared with the reference $^{18}$F-DOPA PET image, and the network parameters are constantly adjusted.

### B. Network Architecture of TC-INN

For better generation, we modify it according to the INN network. Besides, a channel replication technique is incorporated into the network to better improve the learning performance of the reversible network used for PET image generation. The proposed network based on this technique is characterized by higher dimensionality compared to traditional reversible networks.

Fig. 4 illustrates the elaborate architecture of TC-INN. It contains multiple invertible blocks, where each invertible block consists of invertible 1×1 convolution and affine coupling layers. The input image is divided into two halves along the channel dimension. $s$, $t$ and $r$ are transformation equal to dense blocks, which consists of eight 2D convolution layers with filter size 3×3. Each layer learns a new set of feature maps from the previous layer. The size of the receptive field for the first four convolutional layers is 1×1, spanning 2, followed by a rectified linear unit (ReLU). The last layer is a 3×3 convolution without ReLU. The purpose of the Leaky ReLU layers is to avoid overfitting to the training set [33] and further increase nonlinearity. This improves the learning ability of invertible network to generate higher-resolution PET images. In the forward process, input image $x_{^{18}F-FDG}$ is transformed to output image $y_{^{18}F-DOPA}$ by a stack of bijective functions $\{f_i\}_{i=0}^{k}$.

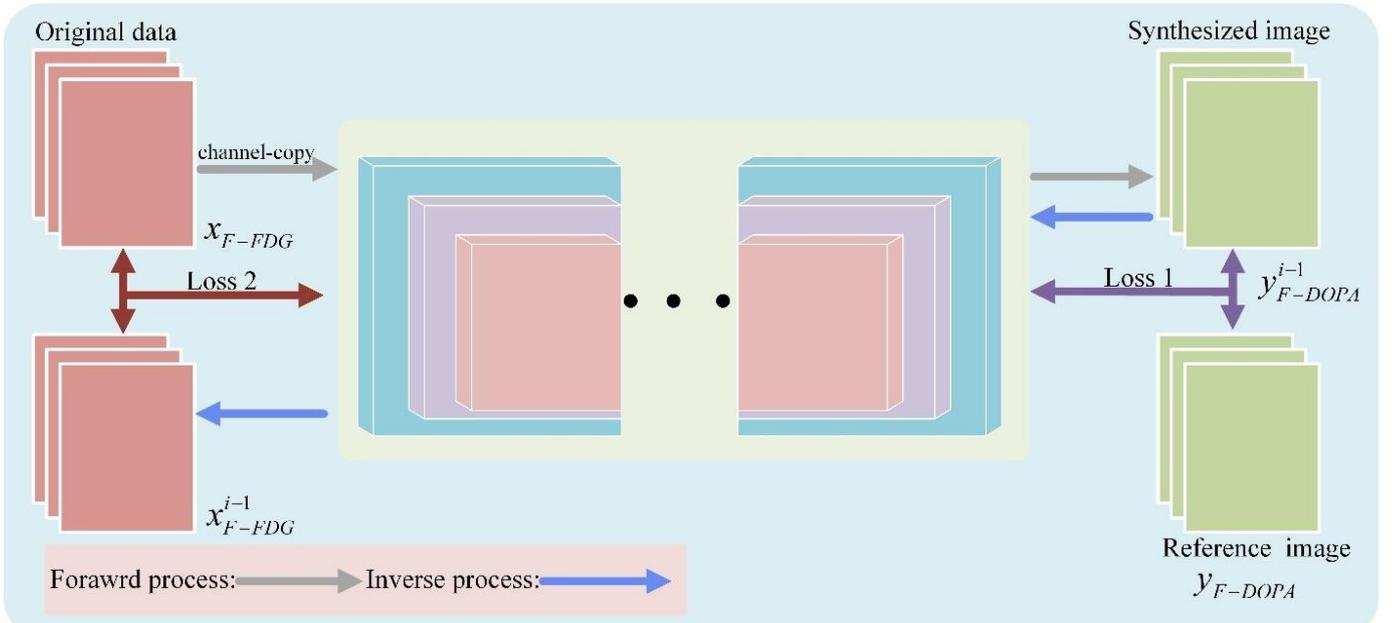

**Fig. 3.** The training process of TC-INN, which provides a reversible connection between the original image and the target image. The forward process of TC-INN produces the synthesized images, and the inverse process aims at recovering the original images.

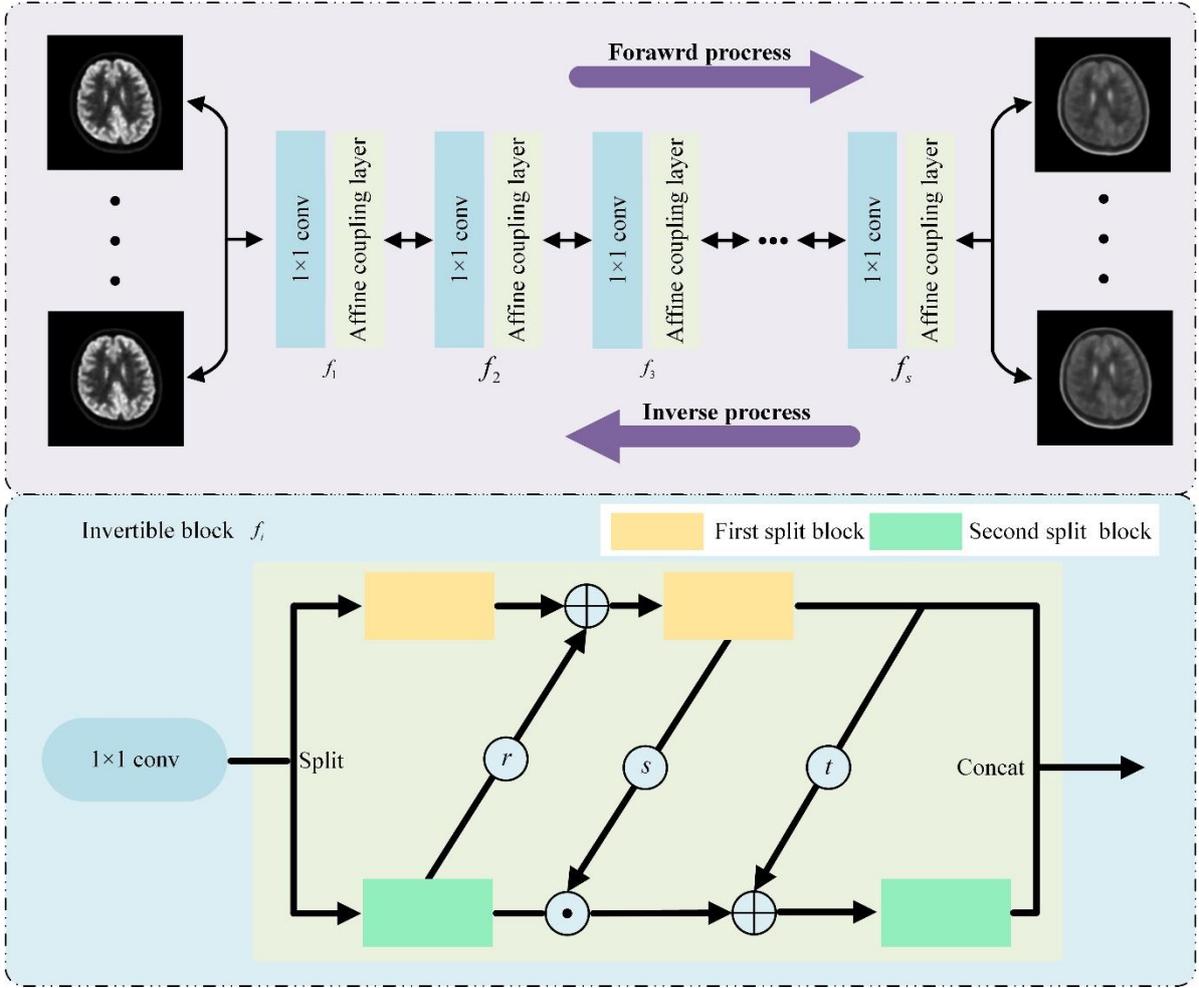

Fig. 4. The pipeline of TC-INN. Each invertible block consists of invertible $1\times 1$ convolution and affine coupling layers. This work illustrates the details of the affine coupling layers on the bottom.

For better generation, the goal of this study is to find a bijective function that maps the data point in the $^{18}$F-FDG PET image $x_{^{18}F-FDG}$ to generated $^{18}$F-DOPA PET image $y_{^{18}F-DOPA}$. Inspired by the core idea in [34], this work adopts an alternative approach that utilizes the affine coupling layers to enable invertibility of individual network. To be specific, the TC-INN designed in this work consists of a stack of reversible, tractable bijection functions $\{f_i\}_{i=0}^{k}$, i.e., $f = f_0 \circ f_1 \circ f_2 \circ \cdots \circ f_k$. For a given observed data sample $x_{^{18}F-FDG}$, we can derive the transformation to $^{18}$F-DOPA PET image $y_{^{18}F-DOPA}$ through the following formula:

$$y_{^{18}F-DOPA} = f_0 \circ f_1 \circ f_2 \circ \cdots \circ f_k(x_{^{18}F-FDG}) \tag{5}$$

$$x_{^{18}F-FDG} = f_k^{-1} \circ f_{k-1}^{-1} \circ \cdots \circ f_0^{-1}(y_{^{18}F-DOPA}) \tag{6}$$

The bijective model $f_i$ is implemented through affine coupling layers. Given a $D$-dimensional input $m$ and $d < D$, the output $n$ in each affine coupling layer is calculated as

$$n_{1:d} = m_{1:d} \tag{7}$$

$$n_{d+1:D} = m_{d+1:D} \odot \exp(s(m_{1:d})) + t(m_{1:d}) \tag{8}$$

where $s$ and $t$ represent the scale and translation functions from $R^D \to R^{D-d}$, and $\odot$ is the Hadamard product. Note that the scale and translation functions are not necessarily invertible. Thus, they need to be implemented through neural networks.

As stated in [34], the coupling layer leaves some input channels unchanged, which greatly restricts the representation learning power of this architecture. To alleviate this deficiency, we firstly enhance the coupling layer by

$$n_{1:d} = m_{1:d} + r(m_{d+1:D}) \tag{9}$$

where $r$ can be arbitrary function from $R^{D-d} \to R^D$. The inverse step is easily obtained by

$$m_{d+1:D} = (n_{d+1:D} - t(n_{1:d})) \odot \exp(-s(n_{1:d})) \tag{10}$$

$$m_{1:d} = n_{1:d} - r(m_{d+1:D}) \tag{11}$$

Next, we utilize the invertible $1\times 1$ convolution presented in [23] as the learnable permutation function to reverse the order of channels for the next affine coupling layer.

### C. Loss Function of TC-INN

As mentioned above, the network relies on continuous improvement of a sequence of invertible blocks that include affine coupling layers, actnorm layers and so on. The accuracy of both invertible blocks depends directly on the design of the corresponding loss function. Many researches leverage multi-component cost function to optimize the network to guarantee the quality of synthesized images. For example, in MM-Synthesis [35], a multi-input multi-output fully convolutional network model for MRI synthesis is trained using a cost function consisting of three cost components. In MM-GAN [24], generation loss and adversarial loss constitute the cost function of this network. The Hi-Net [25] network adds a synthesized loss on the basis of the generation and discriminator loss. Similarly, FGEAN [30] contains not only intra-pixel intensity loss but also gradient information loss.

They all utilize multiple loss functions to ensure the network architecture works well in a variety of synthesis tasks.

Based on the unique invertible nature of the network, we use the euclidean loss to generate representative features. The loss function minimizes the mean squared error between pixel values of the input and the synthesis and fusion image. The training objective of TC-INN is as follows:

$$\mathcal{L}_{hold} = \lambda \mathcal{L}_1 + \mathcal{L}_2 = \lambda \left\| f(x_{^{18}F-FDG}) - y_{^{18}F-DOPA} \right\|_2 + \left\| f^{-1}(y_{^{18}F-DOPA}) - x_{^{18}F-FDG} \right\|_2 \quad (12)$$

where $y_{^{18}F-DOPA}$ is the reference $^{18}$F-DOPA, $f(x_{^{18}F-FDG})$ is the output image from the source image $^{18}$F-DOPA by invertible network $f(\bullet)$, and $\left\| \bullet \right\|_2$ represents the $\mathcal{L}_2$-norm. $\mathcal{L}_1$ stands for the loss function between the synthetic $^{18}$F-DOPA image and reference $^{18}$F-DOPA image. $\mathcal{L}_2$ stands for the loss function between the invertible output image and $^{18}$F-FDG image. Hyper-parameter $\lambda$ is used to balance the two loss functions.

## IV. EXPERIMENTS

### A. Experiment Setup

**1) Datasets.** The Department of Neurology at the First Affiliated Hospital of Sun Yat-sen University recruited a sequence of patients diagnosed with idiopathic Parkinson's disease (PD) between 2020 and 2023 according to the British Parkinson's Disease Association brain bank criteria. The research excluded patients with medical history of dementia, stroke, abnormal brain structure, encephalitis, poor responses (<30%) to levodopa, disease duration of less than 5 years, or were unable to adhere to the study protocol. For the training phase, we used the First Affiliated Hospital of Sun Yat-sen University dataset for generation. The testing dataset was obtained based on an all-digital PET system (Brain PET B320 platform provided by RAYSOLUTION Healthcare Co., Ltd.). In the experiment, we selected 2740 layers of brain PET images chose from 196 brain PET scans for testing. In the experimental section, we show the generated results of 8 images. It should be noted that the study was conducted with the consent of the local ethics committee and the informed consent of the patients.

**2) Data Preprocessing.** First, we cut the image into $200 \times 200$ to retain the desired details and remove unnecessary data parts. Then we apply the data normalization strategy on each pixel of all images. Besides, the quantitative index of PET is standard intake value (SUV), expressed as $Bq/ml$, which is usually greater than 1. Therefore, we linearly scale the raw data values to [0,1] to avoid gradient explosion in deep learning.

**3) Evaluation Metrics Image Registration.** The reason for image registration is that PET images of the same site have different angles and positions under the effect of different tracers. And the structure of the input and output images of the reversible network should be consistent. Therefore, registration of $^{18}$F-DOPA PET and $^{18}$F-FDG PET images that priors to training is necessary to ensure the validity of TC-INN model.

This work utilizes a new end-to-end mechanism for image registration [28] after data preprocessing. Specific, this method presents a discriminator-free image-to-image (I2I) translation mode to replace the original GAN-based I2I mode for accurate registration. In addition, a contrastive PatchNCE loss is designed as a shape-preserving constraint in the translation-based registration model. Local and global alignment loss is also presented to further improve registration performance. The structure of this end-to-end network model is demonstrated in Fig. 5.

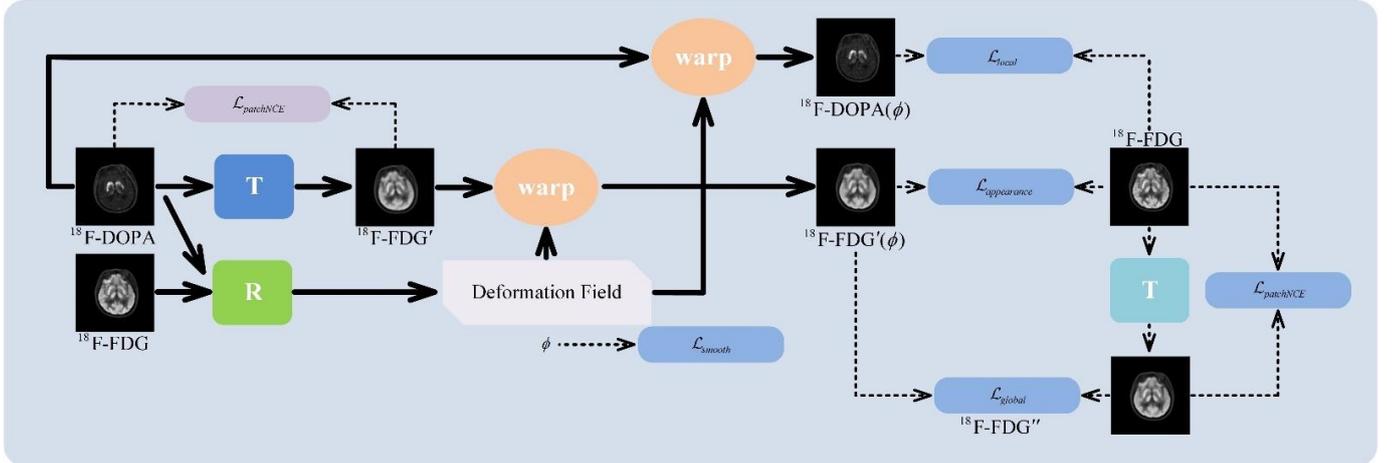

**Fig. 5.** An overview of the end-to-end method for image registration. Specific, this method presents a discriminator-free I2I translation mode to replace the original GAN-based I2I mode and achieve accurate registration.

We denote the registration network as $R$ and the translation network as $T$, as depicted in Fig. 5. Given an image pair ($^{18}$F-DOPA, $^{18}$F-FDG) as the network input, $R$ learns to predict a deformation field $\phi$, which describes how to non-rigidly align $^{18}$F-DOPA to $^{18}$F-FDG. Meanwhile, $T$ takes $^{18}$F-DOPA and $^{18}$F-FDG as the inputs and outputs target-modality images $^{18}$F-FDG′ and $^{18}$F-FDG″, where $^{18}$F-FDG′= $T$ ($^{18}$F-DOPA) and $^{18}$F-FDG″= $T$ ($^{18}$F-FDG′). $^{18}$F-FDG′ is the translated image and $^{18}$F-FDG″ is the reconstructed image of $^{18}$F-FDG. The PatchNCE loss $\mathcal{L}_{patchNCE}$ is employed to force $^{18}$F-FDG′ and $^{18}$F-FDG″ to keep the original structure of $^{18}$F-DOPA and $^{18}$F-FDG respectively. $^{18}$F-DOPA($\phi$) and $^{18}$F-FDG′($\phi$) are warped images from the source image $^{18}$F-DOPA and the translated image $^{18}$F-FDG′. The pixel loss $\mathcal{L}_{appearance}$ enables the appearance transfer of network $T$ between $^{18}$F-FDG′($\phi$) and $^{18}$F-FDG. In addition, local alignment loss $\mathcal{L}_{local}$ and global alignment loss $\mathcal{L}_{global}$ are employed to further improve the registration perfor-mance. For readers interested in delving into the details, a wealth of information can be found in the literature [28].

**4) *Model Training.*** All networks are trained using the Adam solver. We conduct 300 epochs to train the proposed model. For the first 50 epochs, the initial learning rate is set to 0.0001. For every 50 epochs, the learning rate is halved. As the number of epochs increases, the learning rate constantly decreases. The trade-off parameter $\lambda$ is set to 1 during training. Training and testing experiments are conducted using a customized version of PyTorch on an Intel i7-6900K CPU and a GeForce Titan XP GPU.

**5) *Quality Metrics.*** In the testing process, both quantitative and qualitative evaluations are performed to evaluate the proposed method TC-INN. We employ two traditional measures, Peak Signal-to-noise Ratio (PSNR) and Structural Similarity Index Method (SSIM), which are widely applied to determine the quality of the images. That are given as:

$$\text{PSNR}(y, \tilde{y}) = 20\log_{10} Max(y)/\|\tilde{y} - y\|_2 \quad (13)$$

$$\text{SSIM}(y, \tilde{y}) = \frac{(2\mu_y\mu_{\tilde{y}} + c_1)(2\sigma_{y\tilde{y}} + c_2)}{(\mu_y^2 + \mu_{\tilde{y}}^2 + c_1)(\sigma_y^2 + \sigma_{\tilde{y}}^2 + c_2)} \quad (14)$$

where $\tilde{y}$ and $y$ denote the generated PET image and the ground-truth, respectively. PSNR is used to calculate the ratio between the maximum possible signal power and the power of the distorting noise that affects the quality of its representation. SSIM is a perception-based model. In this approach, image degradation is considered as a change in the perception of structural information. RMSE is relatively easy to compute as the average of the absolute difference and difference in quadrature, which is calculated as:

$$\text{RMSE}(y, \tilde{y}) = \sqrt{\sum_1^n (y - \tilde{y})/n} \quad (15)$$

Higher values of PSNR and SSIM indicate better generation of synthesized images. In terms of RMSE, lower values indicate higher prediction accuracy for synthetic DOPA PET images. In order to quantify the performance of the presented method, this work also calculates and compares the absolute mean error (MAE) between synthetic DOPA and reference DOPA brain regions.

$$\text{MAE}(y, \tilde{y}) = \frac{1}{n}\sum_1^n |y - \tilde{y}|/y \quad (16)$$

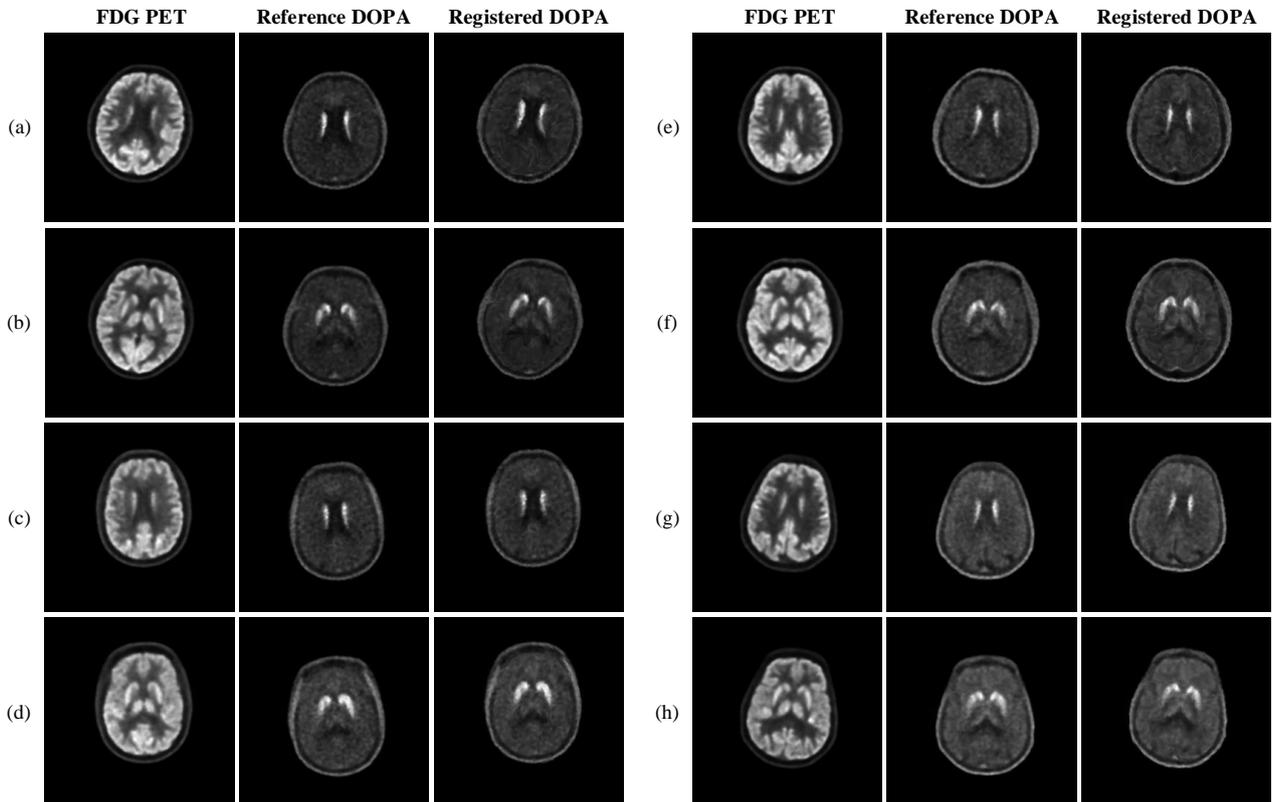

**Fig. 6.** The registration process of $^{18}$F-DOPA. From left to right, the three columns show the $^{18}$F-FDG for reference; the input $^{18}$F-DOPA PET; and the end-to-end network generated registration DOPA diagram.

*B. Registration Experiment*

According to the properties of invertible networks, it is essential to register images before training. Inspired by the literature [28], we use its end-to-end network for PET image registration, and the registration results are indicated in Fig. 6. In the experiment, we choose to display the registration results of 8 images, which implies the effectiveness of the registration network. It also lays a foundation for effective training of invertible model.

As revealed in Fig. 6, this is a registration process in which $^{18}$F-DOPA PET is aligned to $^{18}$F-FDG PET using $^{18}$F-FDG PET as the substrate. The left column is the $^{18}$F-FDG for reference, and the middle column is the input $^{18}$F-DOPA PET. After passing through the network, it is possible to generate the rightmost column, the registered and aligned $^{18}$F-DOPA PET. As can be seen from the figure, the obtained contours are basically consistent with the original input FDG PET, and the details of the intermediate lesion have been preserved, indicating that the image calibration is successful.

*C. PET Image Generation*

To confirm the possibility and effectiveness of converting

FDG to DOPA, we qualitatively and quantitatively demonstrate the generation results of the proposed method. Fig. 7 records the experimental results for 3-channels. The first column on the left is the input, the FDG-PET image. The second column is the desired target, the DOPA-PET image. The third column results are the test results, the generated DOPA-PET images. The generated results contain clear areas of white dots, and the white dot area in the middle of the picture can be considered as the lesion area, which we would like to keep as a criterion for assessing the disease in the experiments. Notably, the overall images are similar to reference DOPA, with some of the original FDG-PET features recaptured in the figure, suggesting that the generated results with a higher signal-to-noise ratio than those in DOPA.

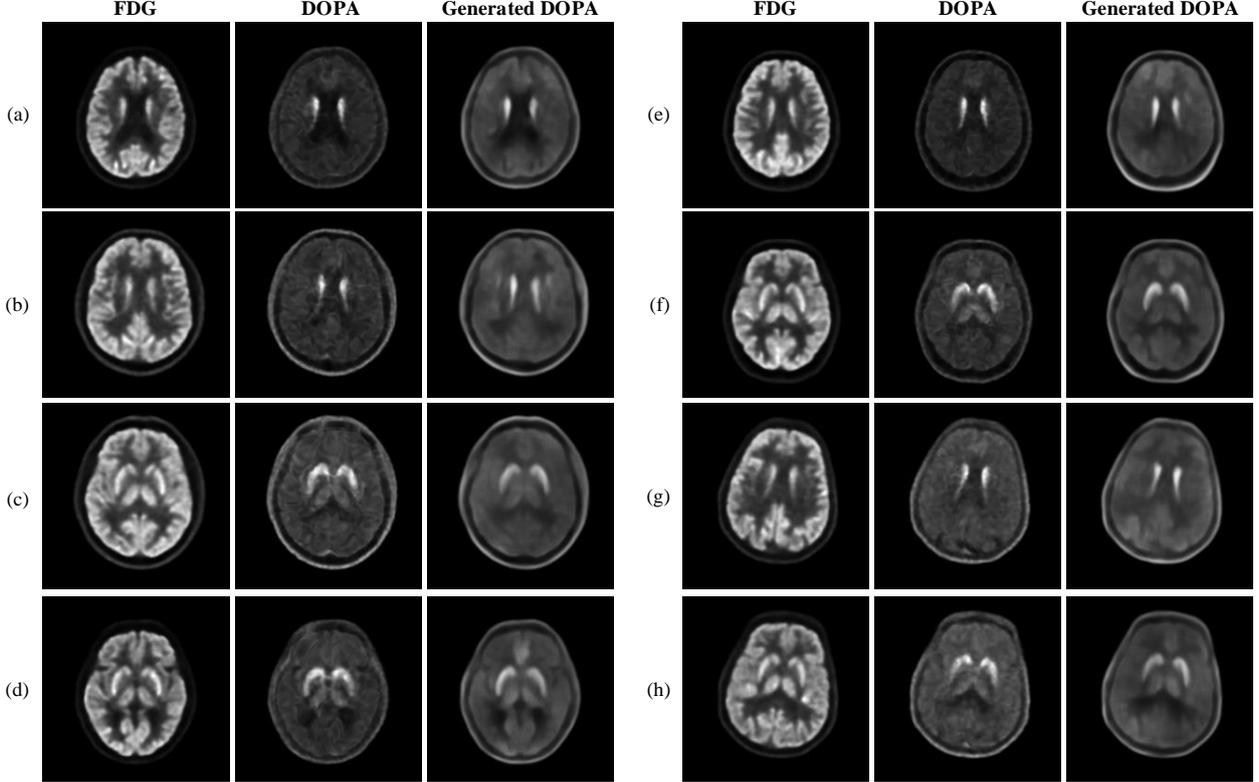

**Fig. 7.** The generation of DOPA with 3-channels. The first column on the left is the input, the FDG-PET image; The second column is the desired target, the DOPA-PET image; The results in the third column are the test results, which are the DOPA-PET images that are generated.

Table I lists the output of the validation metrics. The MAE and RMSE of brain synthesis DOPA images are about 2.72% and 5.34% under the 3-channels condition. The strength similarity between synthetic DOPA and reference DOPA is satisfactory. In addition, the PSNR of the proposed method TC-INN is 25.65 dB. It is worth noting that we plot the training loss function curve for 3-channels in Fig. 8. It can be seen that the loss function converges gradually with the increase of the number of iteration steps. It converges to a stable point at about 100 steps, which proves the superiority of the proposed method.

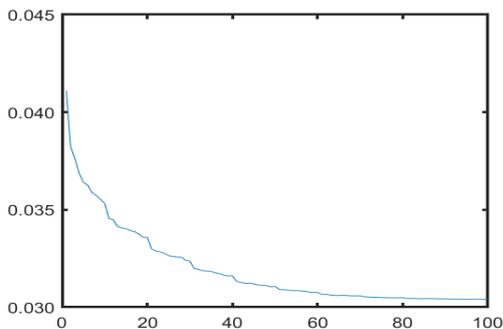

**Fig. 8.** Training loss function curve for 3-channels hypothalamic synthetic DOPA images.

In order to meet the criteria of medical application and to make this work more broadly applicable, more information displayed in Fig. 9. That illustrates the results of generating the problematic image of brain area problem, which shows more detail in generation compared to Fig. 7. This can be visualized by the fact that generated results reflect the detailed features of the original DOPA image.

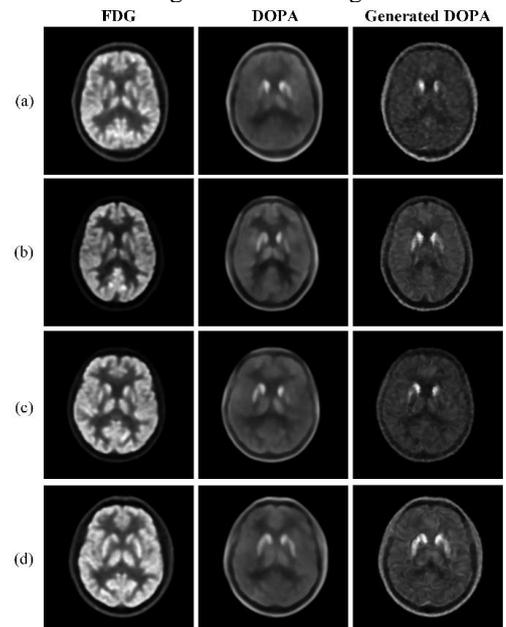

**Fig. 9.** Results of generating DOPA images of brain region issue with striatal defects.

## TABLE I
METRICS VALIDATION AND QUANTITATIVE DEMONSTRATION OF THE PROPOSED METHOD UNDER DIFFERENT CHANNEL NUMBERS.

| Number of channels | PSNR | SSIM | RMSE | MAE |
|---|---|---|---|---|
| 3-channels | 25.645 | 0.8732 | 5.34% | 2.72% |
| 6-channels | 25.890 | 0.8744 | 5.20% | 2.57% |
| 9-channels | 25.965 | 0.8768 | 5.20% | 2.59% |

### D. Ablation Experiment

In this subsection, we verify the influence of the number of channels on the experimental results. From Table I, it can be found that there is no significant difference between the results produced by different channel numbers. For example, the PSNR for a 3-channels result is 25.65 dB, and that for a 6-channels result is 25.89 dB, with a difference of only 0.25 dB. The MAE values of the 3-channels and 9-channels results are only 0.13% different. Fig. 10 shows the qualitatively influence of different channel numbers on the generated results. From the visual perspective, the number of channels has no significant effect on the generated results. Compared to the 3-channels training loss function, Fig. 11 indicates that the 6-channels and 9-channels have better convergence performance. In conclusion, the number of different channels has no effect on the generated results in general, but only differences in the details. It is worth noting that the number of channels has some effects on convergence.

## V. DISCUSSION

To better investigate the value of the experimental results in medical applications, the standardized take values (SUV) of the volume of interest (VOI) are calculated using the following formula:

$$SUV = A_{VOI}/(ID/W) \quad (17)$$

where $A_{VOI}$ is the measured activity in the VOI (in $\mu Ci/mL$), $ID$ is the injected dose (in $mCi$), and $W$ is the body weight of the patient(in $kg$) [37]. In the experiment, corpus striatum is chosen to be the VOI.

First, the generated DOPA-PET images and the original DOPA-PET images are both registered on the AAL template. Then, the area of the corpus striatum is extracted from the image. The VOI allows for the derivation of the average SUV value. Table II tabulates the results for ten patients. From the results, it can be seen that the SUV values of ROIs in the generated images are very close to the SUV values in the original data. This indicates that the proposed method helps in PET dual diagnosis.

Furthermore, it can also be seen from the calculation process that there is still considerable scope for improvement in the measure of this work. For example, optimizing the normalization measures in the data processing stage before the experiment allows the data used in the experiment to be more perfectly restored back to the original SUV. This can be more conducive to the accuracy of the simulation experiment.

## TABLE II
THE CALCULATION RESULTS OF THE AVERAGE SUV ON 710 SETS OF BRAIN PET IMAGES FROM 10 PATIENTS.

| Patient ID | Patient 1 | Patient 2 | Patient 3 | Patient 4 | Patient 5 | Patient 6 | Patient 7 | Patient 8 | Patient 9 | Patient 10 |
|---|---|---|---|---|---|---|---|---|---|---|
| Average SUV in result image | 0.6147 | 0.6226 | 0.7117 | 1.097 | 0.8666 | 0.8637 | 1.725 | 0.5865 | 1.0518 | 0.8777 |
| Average SUV in original image | 0.6319 | 0.6676 | 0.7365 | 1.1264 | 0.8869 | 0.8784 | 1.7552 | 0.5963 | 1.0919 | 0.8833 |
| Average SUV in original image | 0.6319 | 0.6676 | 0.7365 | 1.1264 | 0.8869 | 0.8784 | 1.7552 | 0.5963 | 1.0919 | 0.8833 |

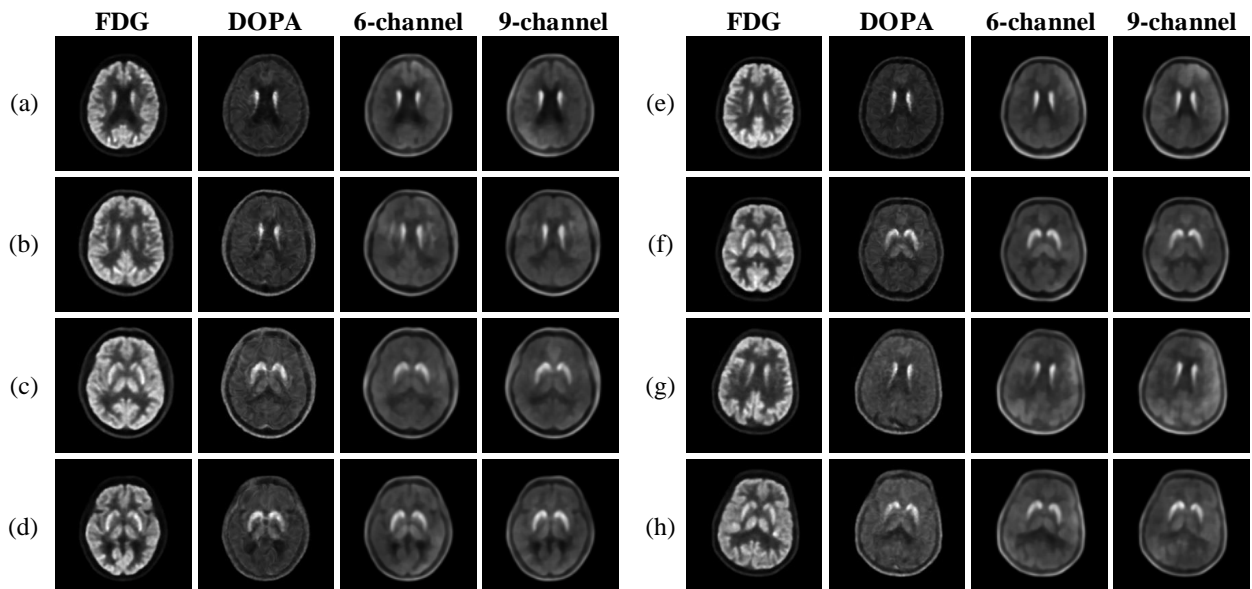

**Fig. 10.** Schematic representation of PET data reconstructed with reference data and synthetic PET images generated, and the results of qualitative effects of different channels on the generated results.

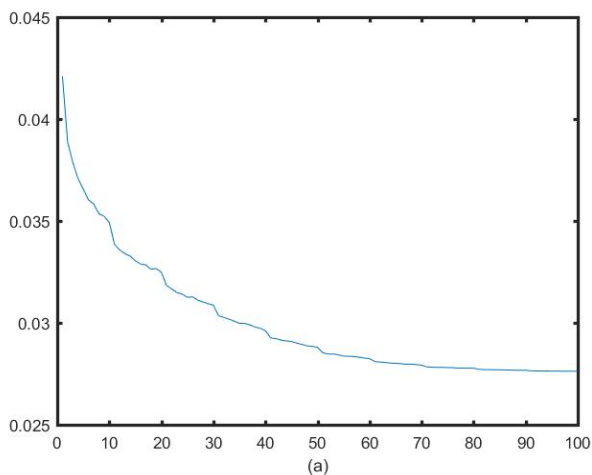 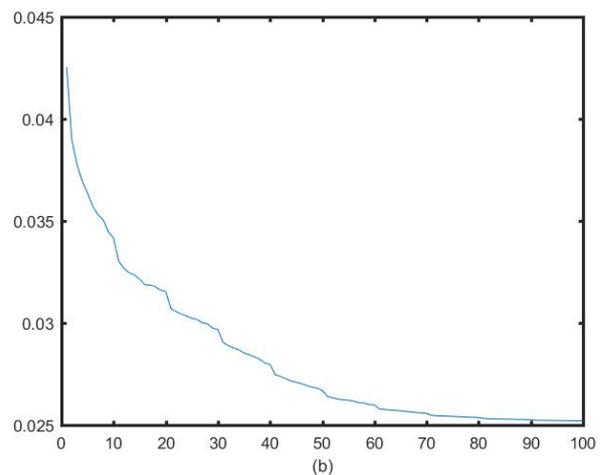

**Fig. 11.** Comparison of the convergence effect of loss functions trained with different channels. Figure (a) on the left is the 6-channels loss function, and figure (b) on the right is the 9-channels training loss function.

## VI. CONCLUSION

Generating hard-to-obtain $^{18}$F-DOPA PET images from ordinary $^{18}$F-FDG PET images by using deep learning method is a promising and innovative technology in the field of PET conversion. This technique does not require the help of other imaging equipment and to some extent satisfactory PET generation results have been obtained. Due to the different imaging angles and focusing areas of different tracers, the imaging angles of different tracers should be registered before network training based on the requirement of network input and output consistency. During network training, this work combines invertible networks with variable enhancement strategies to generate $^{18}$F-DOPA PET images by mining a priori information directly from $^{18}$F-FDG PET images. At the same time, full digital PET reduces the radiation dose in the process of data acquisition and improve the scanning speed. Experimental results showed that the TC-INN method demonstrated satisfactory results in both quantitative and qualitative aspects. In the absence of multiple tracers, the TC-INN method has great potential for PET projection imaging. In forthcoming studies, we will build on the variable augmented invertible network to further investigate the potential of invertible mechanisms in accelerating multimodal imaging and applications of multiple tracer conversion.